\pdfoutput=1
\documentclass{article}
\PassOptionsToPackage{numbers,sort&compress}{natbib}
\usepackage[final,dblblindworkshop]{neurips_2026}
\workshoptitle{Interpreting Agent Behavior (IAB)}
\usepackage[T1]{fontenc}
\usepackage{graphicx}
\usepackage{booktabs}
\usepackage{array}
\usepackage{longtable}
\usepackage[expansion=false]{microtype}
\usepackage{xcolor}
\usepackage{hyperref}
\hypersetup{colorlinks=true,linkcolor=blue!55!black,citecolor=blue!55!black,urlcolor=blue!55!black,pdfauthor={Egor Pakhomov, Erik Nijkamp},pdftitle={},pdfcreator={},pdfproducer={}}

\newcolumntype{L}[1]{>{\raggedright\arraybackslash}p{#1}}

\title{Does an Agent's History Tell You When Compaction Will Hurt? A Modest, Bounded Effect on the TRACE Paired-Replay Corpus}
\author{%
  Egor Pakhomov \\
  Salesforce AI Research \\
  \texttt{epakhomov@salesforce.com} \\
  \And
  Erik Nijkamp \\
  Salesforce AI Research \\
  \texttt{erik.nijkamp@salesforce.com} \\
}

\begin{document}
\maketitle

\begin{abstract}
Many long-horizon agents compact their context on a global rule, usually a token budget, blind to what the agent was doing.
We ask whether the agent's recent behaviour predicts when a compaction will hurt.
TRACE's public corpus of 590 harness-triggered AppWorld compaction boundaries replays each boundary from a re-executed prefix state under the pre-compaction context and under the summary, and records the burden of the next actions: calls that error or repeat a call already made.
We find that pre-boundary history predicts post-compaction harm only weakly.
An internally prespecified contrast by prefix placement is a wide null, and the naive ``has-written'' label behind it turns out to measure trajectory phase.
The best extension-protocol trigger reaches held-out AUROC 0.66 (0.64 on the replicate's own label) against a same-boundary replicate of 0.72; the best frozen, interpretable trigger avoids 21\% of harmful (positive-burden) boundaries while keeping 84\% of compaction opportunities, and exceeds the random-rule expectation on count but not on burden mass (a post hoc comparison).
Whether the best trigger beats a token-budget rule at matched retention cannot be evaluated on the release.
We state what corpora should ship to answer it.
\end{abstract}

\section{Introduction}\label{sec:intro}

Many long-horizon agent harnesses compact \citep{packer2023memgpt,kang2026acon,wu2025resum,desai2026swemarathon,li2026compactionrl}: past a budget, the raw history is replaced by a generated summary \citep{wang2023recursively}, and the decision of \emph{when} is a global rule, blind to what the agent was doing (in the coding-agent scaffolds surveyed by \citet{rombaut2026scaffold}, typically a token threshold or a structural cap)---the natural next step is to defer compaction where the recent history says it is about to hurt.
Agent-invoked compression \citep{liu2026cat}, RL-learned context editing \citep{zhang2025memact} and learned folding \citep{sun2025contextfolding} condition the timing on the history but are judged by task success, which cannot say whether a given compaction hurt; that needs a counterfactual outcome per boundary.

TRACE \citep{min2026trace,trace_software_2026} supplies it (\S\ref{sec:corpus}).
The paired design, the burden channels, the first/later stratification and the thesis that compaction should be judged at the boundary by the agent's next actions, not only by task success (process-level evaluation \citep{ma2024agentboard}; failure attribution \citep{zhang2025whichagent}; summaries can lengthen trajectories by masking failure signals \citep{lindenbauer2025complexitytrap}) are the source's \citep{min2026trace,trace_software_2026}.
Behavioural interpretation of agents mostly labels completed transcripts \citep{gao2026actonomy}; a paired replay is the intervention such a science calls for \citep{chen2025behavsci}.
We use the released cohort as is and ask: does what the agent did before a boundary predict whether compaction there will hurt?

\textbf{Contributions.} (i) An internally prespecified placement test with a post hoc diagnostic of what a naive write label measures (\S\ref{sec:placement}). (ii) A task-held-out risk ranking of boundaries and an opportunity-retention frontier against global ON/OFF and the random-rule expectation at matched retention (retention: the share of opportunities a rule still allows; the random-rule expectation: what a uniformly random rule achieves), not reported by the source \citep{min2026trace} (\S\ref{sec:trigger}). (iii) A benchmark of two fixed learners against a split-half replicate yardstick under a frozen extension protocol (post-freeze additions dated in App.~\ref{app:prereg}).
The placement contrast is a wide null. The seven Table~\ref{tab:frontier} policies sit two to seven points above the random-rule expectation on count (only the stump's interval crosses zero; the two-feature logistic does so too when refit under stem-grouped folds, App.~\ref{app:ext}, Table~\ref{tab:ext2c}) and zero to eleven points on burden mass, clear of zero under replay resampling for three of the seven (App.~\ref{app:policy}, \ref{app:ext}). Held-out AUROC \citep{hanley1982meaning} is 0.404--0.635 across Table~\ref{tab:auroc}'s original-label cells for the interpretable policies (0.55--0.62 on the primary target) and 0.66 for the boosted ones, against 0.72 for a replicate of the same boundary.
Interpretability lessons: the phase proxy (\S\ref{sec:placement}) and first-compaction fragility (\S\ref{sec:discussion}).

\section{Corpus and measurement}\label{sec:corpus}

\textbf{Cohort.}
TRACE's frozen cohort holds 590 compaction boundaries from 275 of 294 trajectories on 147 AppWorld tasks \citep{trivedi2024appworld} (140 with a boundary; App.~\ref{app:prereg}).
Agent and compressor are one model (minimax-m3; 4,096-token window) \citep[\S3]{min2026trace}; boundaries are harness-triggered, never agent-chosen; 275 are first compactions, 315 later.
At each boundary the recorded code prefix is re-executed into a fresh environment to rebuild the state (the replay is identified by a hash of task, boundary step and prefix code, not of the reached state; deltas inherit any restoration variance) and the agent runs at most five actions from the PRE rendering (full raw prefix at first compactions; previous summary plus raw suffix later) or the POST rendering (recorded summary plus retained last turn); five samples per arm at first compactions, three at later ones \citep[\S3.3]{min2026trace}.
\textbf{Channels.}
Blocked, refetch and union channels are defined in App.~\ref{app:prereg}.
\textbf{One boundary.}
The boundary comes from task \texttt{6ea6792\_2} (easy), a first compaction after seven steps, chosen for clarity (dU@3 1.8, five times the cohort mean of 0.35). Its prefix function set is \texttt{api\_docs.show\_api\_descriptions}, \texttt{api\_docs.show\_api\_doc}, \texttt{supervisor.show\_account\_passwords}, \texttt{venmo.login} and \texttt{venmo.show\_received\_payment\_requests}; its last observation is a page of pending payment requests.
Under PRE (the full raw prefix) all five rollouts request page 2 (empty), then print the requests already held in Python variables or keep browsing; none of the first three actions is blocked or a refetch.
Under POST (the summary plus the retained turn) all five request page 2; four then hit a \texttt{NameError} because the variables the summary describes are no longer bound (blocked) and re-issue the prefix call \texttt{show\_received\_payment\_requests(page\_index=0)} (refetch); the fifth re-reads the Venmo API list it had already fetched (refetch). The boundary's dU@3 is 1.8 (PRE 0.0, POST 1.8; blocked 0.8, refetch 1.0).
For each boundary and horizon $k\in\{1,3,5\}$, each rollout contributes its count over the first $\min(k, L)$ actions, $L$ its executed actions; the paired delta is mean(POST)\,$-$\,mean(PRE) of that count, giving dE (blocked), dR (refetch) and dU (union); exposure and completion by arm are in App.~\ref{app:cells}.
dU at $k=3$ is the preregistered primary outcome; 344 of 590 boundaries have dU@3\,$>0$.
\textbf{Statistics.}
Unless stated otherwise, every interval is a 95\% percentile bootstrap over tasks as clusters \citep{cameron2008bootstrap,efron1993bootstrap} with fixed seeds, 10,000 draws for the preregistered analyses and 2,000 for the extensions (App.~\ref{app:ext}).

\textbf{What the release exposes before a boundary.}
The release exposes sorted call and function sets, prefix step count, compaction index, task identifier and difficulty, and the last three observations---not the ordered action sequence, token counts, compression ratio or summary text (App.~\ref{app:prereg}).

\section{Is burden localised by pre-boundary placement?}\label{sec:placement}

\textbf{Label.}
Each AppWorld endpoint carries an HTTP method in its decorator (GET = \emph{read}; POST/PUT/PATCH/DELETE = \emph{write}); the map covers all 119 functions observed (App.~\ref{app:map}), so every boundary is labelled.
The map was checked by an LLM-agent-authored table encoded in the analysis (43/43 non-GET functions) and, in a second audit, by two labelers of different model families given the READ/WRITE and SESSION/SUBSTANTIVE definitions but not the map (119/119; relabelling moves no boundary; App.~\ref{app:map}).
Cells: 494 boundaries with a write in the prefix, 96 without (read-only).
\textbf{Prespecified contrast.}
dU@3, post-write minus read-only: $-0.062$ [$-0.209$, $+0.079$] (any prior write $+0.335$ [$+0.269$, $+0.405$] over 494 boundaries / 135 tasks); stem-clustered, post hoc sensitivity: [$-0.238$, $+0.100$] (App.~\ref{app:policy}).
The frozen criterion for an informative answer (interval excluding zero, or half-width at most 0.10) was not met (half-width 0.144; criterion (b) was out of reach at the realized cell sizes, computed after the fact from the observed SDs, App.~\ref{app:prereg}).

\begin{table}[t]
\centering
\footnotesize
\setlength{\tabcolsep}{4pt}
\caption{Boundaries by pre-boundary function set; the no-write row is the frozen two-cell label's read-only cell and the substantive-write row is the post hoc label's cell, both with frozen-seed intervals (App.~\ref{app:cells}); the login-only interval is derived (seed 20260950); login-only = writes only to session endpoints (App.~\ref{app:map}). Frozen two-cell label and prespecified contrast: \S\ref{sec:placement}.}
\label{tab:label}
\begin{tabular}{@{}lrrrrl@{}}
\toprule
Pre-boundary function set & Boundaries & Tasks & Prefix steps & Compaction index & dU@3 [95\% CI] \\
\midrule
No write & 96 & 60 & 7.9 & 1.22 & $+0.398$ [$+0.266$, $+0.533$] \\
Login-only write & 368 & 123 & 15.1 & 1.89 & $+0.391$ [$+0.313$, $+0.472$] \\
Substantive write & 126 & 64 & 22.1 & 3.14 & $+0.172$ [$+0.048$, $+0.291$] \\
\bottomrule
\end{tabular}
\end{table}

\textbf{What the label measures.}
A post hoc diagnostic (App.~\ref{app:prereg}): of the 494 ``write-prefixed'' boundaries, 368 have no write beyond login or session calls; 126 contain a state change to task-relevant application data.
Here ``has-written'' is mostly ``has been running for a while'': a trajectory-phase proxy.
\textbf{Post hoc relabelling.}
Excluding the login/session endpoints from the write set is outcome-free but was defined after the original-label results had been seen, so it is post hoc and carries no gate.
It yields an uncontrolled contrast of $-0.22$ that falls to $-0.004$ [$-0.19$, $+0.18$] once task fixed effects, log prefix steps and a first-compaction indicator are added (App.~\ref{app:cells}): the apparent placement effect is consistent with confounding by the measured phase variables.

\section{Can a history-based trigger beat random selection at matched retention?}\label{sec:trigger}

\textbf{Policies.}
A trigger policy scores each boundary from pre-boundary observables and blocks compaction at or above a threshold.
Three interpretable classes: a stump and a depth-2 tree over six observables (App.~\ref{app:policy}), and a logistic regression on two: has-written and recent errors.
Target: $1[\mathrm{dU@3}>0]$; evaluation task-held-out five-fold, all scores out-of-fold.
\textbf{Offline value.}
An allowed boundary incurs the POST arm; a blocked one keeps the pre-compaction rendering and incurs the PRE arm; both arms are observed for every boundary, so a policy's value is a plug-in sum over logged arms with no new rollouts, not an off-policy estimate \citep{dudik2011doubly}.
The x-axis is the fraction of compaction opportunities retained (no token counts are released, so opportunities are equal-weight; App.~\ref{app:prereg}); the y-axis is the fraction of positive-burden boundaries blocked; the random-rule expectation is the diagonal $y=1-x$ (analogous to a risk--coverage curve \citep{geifman2017selective,elyaniv2010foundations}).

\begin{figure}[t]
\begin{minipage}[t]{0.23\linewidth}
\vspace{0pt}
\centering
\IfFileExists{figures/fig1-frontier-landscape-v3.pdf}{\includegraphics[width=\linewidth]{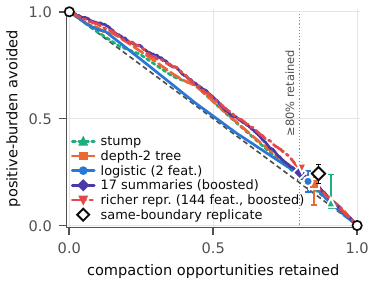}}{\IfFileExists{figures/fig1-frontier-landscape-v2.pdf}{\includegraphics[width=\linewidth]{figures/fig1-frontier-landscape-v2.pdf}}{\IfFileExists{figures/fig1-frontier-landscape.pdf}{\includegraphics[width=\linewidth]{figures/fig1-frontier-landscape.pdf}}{\IfFileExists{figures/fig1-frontier-compact.pdf}{\includegraphics[width=\linewidth]{figures/fig1-frontier-compact.pdf}}{\IfFileExists{figures/fig1-frontier.pdf}{\includegraphics[width=\linewidth]{figures/fig1-frontier.pdf}}{\fbox{\parbox[c][2.0in][c]{0.94\linewidth}{\centering \textbf{[FIG1]}}}}}}}}
\end{minipage}\hfill
\begin{minipage}[t]{0.75\linewidth}
\vspace{0pt}
\caption{Out-of-fold frontier (task-held-out folds for the three frozen policies; stem-grouped folds averaged over 20 repetitions for the two boosted ones; Table~\ref{tab:frontier}'s two lower logistic rows are not drawn): positive-burden boundaries (dU@3\,$>0$; 344 of 590) blocked vs.\ equal-weight compaction opportunities retained; diagonal = random-rule expectation; hollow circles = global ON $(1,0)$ / OFF $(0,1)$; markers with 95\% intervals at $\ge 80\%$ retained (Table~\ref{tab:frontier}). Hollow diamond: the same-boundary replicate used as a trigger score (a yardstick, not a policy; App.~\ref{app:ext}). (Full width: App.~\ref{app:policy}.)}
\label{fig:frontier}
\end{minipage}
\vspace{2pt}
\makeatletter\def\@captype{table}\makeatother
\caption{Operating points at $\ge 80\%$ retained. $\Delta$ = paired advantage over the random-rule expectation $1-$Ret.\ (a fixed expectation, not a realized draw); replicate row scored against half-sample labels. Rows are read at their own operating points (Ret.) under different fold schemes (task-held-out for the frozen policies; stem-grouped $\times$ 20 repetitions for the extension models), so rows are not directly comparable.}
\label{tab:frontier}
\centering
\footnotesize
\setlength{\tabcolsep}{4pt}
\begin{tabular}{@{}llrll@{}}
\toprule
Policy & Folds & Ret. & Avoided [95\% CI] & $\Delta$ vs random [95\% CI] \\
\midrule
Stump & task & 0.908 & 0.108 [0.083, 0.238] & $+0.016$ [$-0.001$, $+0.048$] \\
Depth-2 tree & task & 0.842 & 0.195 [0.096, 0.246] & $+0.037$ [$+0.012$, $+0.064$] \\
Logistic (2 features) & task & 0.837 & 0.206 [0.156, 0.256] & $+0.044$ [$+0.017$, $+0.070$] \\
\midrule
17 engineered summaries (boosted) & stem & 0.800 & 0.247 [0.217, 0.277] & $+0.047$ [$+0.019$, $+0.079$] \\
Richer representation (144 features, boosted) & stem & 0.800 & 0.267 [0.238, 0.290] & $+0.067$ [$+0.040$, $+0.092$] \\
Logistic, 144 features & stem & 0.800 & 0.273 [0.243, 0.296] & $+0.073$ [$+0.045$, $+0.098$] \\
Logistic, summaries + difficulty & stem & 0.800 & 0.273 [0.243, 0.299] & $+0.073$ [$+0.045$, $+0.100$] \\
\midrule
Replicate (yardstick) & half-sample & 0.866 & 0.243 [0.198, 0.286] & $+0.109$ [$+0.082$, $+0.134$] \\
\bottomrule
\end{tabular}
\end{figure}

\textbf{Result} (Fig.~\ref{fig:frontier}, Table~\ref{tab:frontier}).
At $\ge 80\%$ retained, the logistic policy avoids 0.206 [0.156, 0.256] of positive-burden boundaries at 0.837 retained (strict threshold; conditional on fitted out-of-fold scores; App.~\ref{app:policy}); the advantage over the random-rule expectation of 0.163 is $+0.044$ [$+0.017$, $+0.070$] (a post hoc interval; App.~\ref{app:prereg}); by burden mass (share of all positive dU@3) the logistic policy has no advantage ($-0.001$ [$-0.034$, $+0.035$]), while the depth-2 tree and the boosted trigger clear zero on the task-only bootstrap with the threshold reselected per draw (Table~\ref{tab:ext2c} for the boosted trigger; App.~\ref{app:policy} for the depth-2 tree).
The frozen gate G4 (App.~\ref{app:prereg}) criterion (half-width $\le 0.15$ at the 80\% point) was met, with half-width 0.050.
\textbf{Engineered summaries of the released observables.}
Under a second protocol (frozen before any extension output; App.~\ref{app:ext}), an L2 logistic regression and a shallow gradient-boosted classifier were given seventeen engineered summaries of the released observables (App.~\ref{app:ext}) under five-fold cross-validation grouped by task stem, the scenario identifier a task shares with its variants (48 stems, 20 repetitions); the logistic cell with the largest operating-point advantage (summaries plus task difficulty, Table~\ref{tab:ext2c}) is also in Table~\ref{tab:frontier}.
The best cell, the boosted classifier on the seventeen history features, reaches held-out AUROC 0.66 [0.61, 0.70] (App.~\ref{app:ext}); at 80\% retained it blocks 0.247 [0.217, 0.277] of positive-burden boundaries and 0.238 [0.201, 0.284] of positive burden mass, against a random-rule expectation of 0.200 ($+0.047$ [$+0.019$, $+0.079$] by count; Table~\ref{tab:frontier}, Fig.~\ref{fig:frontier}); these extension intervals are unadjusted for selection and multiplicity (App.~\ref{app:policy}, \ref{app:ext}).
\textbf{Richer representations.}
Adding one indicator per prior function and app (144 features) gives held-out AUROC 0.66 [0.61, 0.71] and $+0.067$ [$+0.040$, $+0.092$] over the random-rule expectation at 80\% retained; hashed observation text or exact calls did not improve held-out AUROC point estimates (0.61--0.65) (full statement in App.~\ref{app:ext3}; both 144-feature models are in Table~\ref{tab:frontier}).
\textbf{Replicate yardstick.}
Splitting each boundary's PRE and POST rollouts into disjoint halves (200 random splits), one half's paired delta predicts the sign of the other's with AUROC 0.69--0.72 at $k=3$ depending on which half supplies the label (0.72 [0.69, 0.75] under the frozen extension protocol's assignment; first 0.74, later 0.68); the half--half correlation is 0.40 and the task-level share of variance is near zero (App.~\ref{app:ext}).
Burden is thus moderately reproducible in ranking terms (AUROC 0.72, $r$ 0.40) and only weakly in sign (three-class agreement 0.55, $\kappa$ 0.28); little of its variation is attributable to task identity in this cohort.
Against this yardstick the three frozen policies (0.55--0.62 at $k=3$) fall short of what a replicate of the same boundary predicts; the extension protocol's overlap rule would read the boosted trigger (0.66 [0.61, 0.70]) as within the replicate's range, but in the post hoc same-label comparison (amendment D) the replicate leads the seventeen-summary trigger by $+0.07$ [$+0.04$, $+0.11$] and the 144-feature one by $+0.06$ [$+0.02$, $+0.10$] (App.~\ref{app:ext}), so on that comparison the best trigger falls short of what a replicate of the same boundary predicts.

\section{Discussion and limitations}\label{sec:discussion}

\textbf{For harness design.}
At a boundary the released observables barely separate harmful compactions from harmless ones, and the evidence does not show a per-boundary trigger a harness designer would adopt on this corpus. For the representations and learners tested, the count advantage over the random rule is two to seven points. Against the same half-sample label, the boosted seventeen-summary trigger's 0.64 recovers about two-thirds of the replicate's 0.72 discrimination above chance (0.14 of 0.22 AUROC points; App.~\ref{app:ext}); with three-class sign reliability of only 0.55 ($\kappa$ 0.28) the target itself is noisy, so the same numbers also support an attenuation reading.
When this agent is fragile: the source's shipped curves already show first compactions carry more burden than later ones; under task fixed effects the gap is $+0.274$ [$+0.084$, $+0.474$], the stump's shorter-history leaf carries the higher positive-burden rate, and the tree's leaf for a short history ending in a near-empty observation reaches 0.89--0.93 in four of five folds on its fitted rows (0.71 in the fifth; 28--36 boundaries per fold, mostly first compactions; App.~\ref{app:policy}).
The interpretability lesson: check a ``has-written'' label against prefix length and compaction index before reading it as placement.
The outcome is a meaningful prediction target---the replicate yardstick shows the burden is reproducible enough---but whether its signal is available in the pre-boundary history is not settled by the models and representations tested here; corpora should ship ordered actions, token counts, summary text and more replays.

\textbf{Limitations.}
One agent model, one compressor and one window (4,096 tokens, deliberately tight); harness-triggered boundaries on AppWorld only, so agent-chosen compaction \citep{liu2026cat} is out of scope; the offline value prices only the next $k$ actions, not a later compaction a blocked one may only postpone; coarse released observables; 140 task clusters, 55 with both placement cells; per-boundary deltas from three to five replays per arm (noisy); frontiers pooled over fold models (5 or 100), not a deployable fitted policy (App.~\ref{app:policy}); no data redistribution (the source carries no licence file)---we cite the release; the analysis record is available on request.
LLM-use statement: App.~\ref{app:llm}.

\begin{ack}
This work was funded by Salesforce.
\end{ack}

\bibliographystyle{unsrtnat}
\bibliography{main}

\appendix

\section{Reproduction and preregistration record}\label{app:prereg}

\textbf{Source.}
TRACE repository at revision \texttt{9be7d669}, cloned once; no re-download and no copy of the data outside the clone.
The corpus documentation states 147 tasks, 294 trajectories, 590 boundaries, 4,641 stored rollouts (4,640 analysed: one rollout is kept with a recorded collection error and skipped by scoring) and 20,133 executed steps; 140 of the 147 tasks and 275 of the 294 trajectories carry at least one boundary; the release's run configuration names the agent model \texttt{minimax-m3}; the source preprint states the same model served as compressor \citep{min2026trace}; window 4,096, matching the source preprint's main experiments (its \S5.1). The source also releases its own annotation of compaction mechanisms on 46 compactions in 14 trajectories per the source's consolidated annotation summary (its data glossary counts the 12 trajectories of the first annotation pass), which we do not use (46 annotated compactions against 590 boundaries).

\textbf{Renderings.}
At a trajectory's first compaction the PRE rendering is the full raw prefix; at later compactions it is the previous real summary plus the raw suffix since it, and no previous summary is ever fabricated. The POST rendering is always the recorded summary plus the single retained raw turn.

\textbf{Released observables.}
Sorted sets of prior calls and functions, prefix step count, compaction index, task identifier and difficulty, and the last three observations, but not the ordered action sequence, token counts, compression ratio or summary text; three preregistered features are therefore unmeasurable and the two-cell collapse is a dated amendment (below).

\textbf{Channels.}
Per executed action, \emph{blocked} means the action returned an error under AppWorld's native error contract (the source's per-step error flag); \emph{refetch} means its exact action signature (function and arguments) had already been observed before the boundary or earlier in the same rollout; \emph{union} means blocked or refetch (this channel sense of \emph{blocked} is distinct from a policy blocking a compaction, \S\ref{sec:trigger}). These are the source's definitions \citep[Fig.~4]{min2026trace}; per rollout each channel is counted over the first $\min(k,L)$ actions (\S\ref{sec:corpus}).

\textbf{Reproduction (gate G1).}
The source ships six scan tables and a test that regenerates them.
Under Python 3.10.8 four of the six regenerated tables were byte-identical to the shipped ones and the two bootstrap-bearing tables differed at the $10^{-15}$ level (Table~\ref{tab:sha}); the cause is float summation order across interpreter versions, not randomness (the source bootstrap is seeded), and under Python 3.12.13 all six were byte-identical and the test passed 6/6.
Downstream analysis used the Python-3.10 fresh scan as its frozen basis; its deterministic tables are byte-identical to the shipped ones, so no number in this paper depends on the difference; the row-level comparison is in the frozen reproduction record.

\begin{table}[htbp]
\centering
\footnotesize
\caption{Shipped scan tables versus a fresh scan (SHA-256, first 16 hex digits).}
\label{tab:sha}
\begin{tabular}{@{}llccc@{}}
\toprule
Table & Shipped SHA-256 & Python 3.10 fresh & Python 3.12 fresh \\
\midrule
\texttt{boundary\_wide.csv} & \texttt{8534faccb3a97dec} & identical & identical \\
\texttt{boundary\_long.csv} & \texttt{34a61f99aa9d7af7} & identical & identical \\
\texttt{step\_level.csv} & \texttt{eba9b5d8075c90ea} & identical & identical \\
\texttt{curves.csv} & \texttt{0cd6cee8d4f5dfda} & differs (\texttt{6fba4773d74415b7}) & identical \\
\texttt{marginal.csv} & \texttt{2d144932ff1bd365} & differs (\texttt{8b2adbb1326b5d98}) & identical \\
\texttt{error\_subtypes.csv} & \texttt{6a2969efd8f15b91} & identical & identical \\
\bottomrule
\end{tabular}
\end{table}

\textbf{Gates as frozen, and their status.}
The criteria were written before any stratified analysis ran and reference execution quality and informativeness only, never the direction of an estimate.
\begin{itemize}\itemsep1pt
\item \textbf{G1 Reproduction.} The source's reproduction test passes and a fresh scan matches the shipped tables by SHA-256. \emph{Met} on Python 3.12 (above).
\item \textbf{G2 Taxonomy coverage.} At least 90\% of the 590 boundaries receive a deterministic read/write label; an audit clause; each analysed cell has at least 30 boundaries, else a prespecified collapse to two cells. Met on coverage (590/590) and cell size (494/96); the audit clause as written (``hand-audit agreement $\ge$ 45/50'', frozen text, verbatim) was not met---no person audited---and was replaced by the dated G2 audit-clause amendment below (a map-blind audit by two LLM labelers, 119/119, App.~\ref{app:audit2}, reported descriptively with no pass threshold); a separate encoded semantic-map check (43/43, App.~\ref{app:map}) preceded it; G2 is met only as amended: the amended audit clause is descriptive (agreement reported, no threshold), so G2 as amended reduces to its coverage and cell-size clauses, both met.
\item \textbf{G3 Informative heterogeneity answer (either direction).} The post-write minus read-only contrast on dU@3 has a task-clustered 95\% interval that (a) excludes zero or (b) has half-width $\le 0.10$; wide and crossing zero means the placement half is not the headline and the paper falls back to the policy half if G4 passes. \emph{Not met}: $-0.062$ [$-0.209$, $+0.079$], half-width 0.144. Criterion (b) was out of reach at the realized cell sizes by a post hoc unclustered normal approximation: the unclustered half-width at the realized cell sizes is already 0.129 (sample SDs 0.557 and 0.728 at $n = 96$ and 494; computed after the fact from the observed SDs), so G3 could only have passed by excluding zero; the null's width is a design consequence, not a finding.
\item \textbf{G4 Informative policy answer (either direction).} Task-held-out frontier estimates for every policy class with bootstrap intervals; the best policy's avoided fraction at the 80\% point has interval half-width $\le 0.15$; beating or losing to global rules is reportable either way. \emph{Met}: the frozen logistic's half-width is 0.050 task-only and 0.052 nested (App.~\ref{app:policy}); the criterion ($\le 0.15$) was met.
\item \textbf{G5 Layer distinctness from the source.} The source has not already published the taxonomy-heterogeneity layer or the offline-trigger layer; the paper claims only clear layers. \emph{Met}: the source preprint \citep{min2026trace} reports the paired PRE/POST design, aggregate post-compaction action-position effects (its \S3.3, Fig.~4), a compression-prompt optimizer driven by summary preferences, and a comparison of fixed context budgets, 16K/8K/4K/2K, under FIFO truncation versus summarisation (its \S3.1--3.2, Figs.~1--2); it reports no boundary-risk trigger, selective-trigger frontier or task-held-out policy evaluation.
\item Write-up proceeds iff G1, G2 and G5 pass and at least one of G3/G4 passes. Satisfied via G4, with G2 as amended; this paper is the policy half, and \S\ref{sec:placement} reports the G3 null as a null.
\end{itemize}

\textbf{Amendments (dated; the originals were never edited).} All freezes were internal: dated files with SHA-256 hashes, available from the authors on request; no third-party registry was used.
\begin{itemize}\itemsep1pt
\item \emph{2026-09-01, G1$'$.} After the Python-3.10 run, the byte-identity requirement was relaxed for the two bootstrap-bearing tables to numerically equivalent point estimates (differences at the $10^{-15}$ level) and statistically consistent intervals, with the fresh scan as the frozen downstream basis and the discrepancy reported here verbatim. \emph{2026-09-02, correction:} the source bootstrap is seeded; the differences are summation order across interpreter versions; Python 3.12 gives byte identity for all six tables. G1 is met on its original letter; the amendment stands as a provenance record.
\item \emph{2026-09-02, G2$'$ (label and observables).} Logged before any amended-label output was generated in the full run. (a) A second, post hoc label excludes session endpoints (name suffix in \{\texttt{login}, \texttt{logout}, \texttt{set\_access\_token}, \texttt{signup}, \texttt{send\_verification\_code}, \texttt{reset\_password}, \texttt{send\_password\_reset\_code}\}) from the write set; the original label and all its outputs are kept unchanged, and every downstream stage is run under both labels with the same seeds. \textbf{The two-cell collapse had already been made in the earlier script, and the original-label results had been computed and seen on 2026-09-01 before the second label was defined on 2026-09-02; that is why it is post hoc and carries no gate.} (b) The four-cell taxonomy collapsed to two cells because \texttt{last\_action\_type}, \texttt{unresolved\_commitment} and \texttt{compression\_ratio} are unmeasurable in the release, not because of cell size; ``any write anywhere in the pre-boundary function set'' is the substitute observable and the task-id stem the app-domain proxy. (c) \texttt{position\_fraction} is dropped from every feature set because it is 1.0 in all 590 rows by construction; the two-feature logistic keeps its artifact name. (d) A first-compaction indicator is added to the controlled regression and a first-compaction-only contrast is added with its own seed.
\item \emph{2026-09-02, G4$'$ (frontier axis).} Logged before any frontier output was read. The release carries no per-boundary token counts, so the frontier x-axis is the fraction of equal-weight compaction opportunities retained; the G4 criterion is unchanged and direction-blind.
\item \emph{2026-09-02, extension protocol.} The replicate yardstick and the engineered-summary trigger (App.~\ref{app:ext}) were frozen, with their interpretation rules, before any extension output was computed.
\item \emph{2026-09-02, G2 audit clause.} The frozen clause named a hand audit that no person performed; a map-blind audit by two LLM labelers of different model families was frozen as a separate mini-protocol at 12:20, before any audit call and with no threshold deciding anything, and run at 12:38 (App.~\ref{app:audit2}); the G2 status above records the substitution.
\item \emph{2026-09-02, advantage over the random rule.} Paired task bootstrap of avoided $-$ $(1-\mathrm{retained})$ at the operating point, for the frozen policies and then for the extension models: a post-freeze addition made during internal pre-submission review, computed 13:00; not preregistered.
\item \emph{2026-09-02, amendments A--F, all direction-blind post-freeze additions made during internal pre-submission review.} A (12:40, nested task $\times$ replay bootstrap) and B (12:40, mixed-label paired replicate $-$ trigger difference): computed 13:02. C (13:52, EXT-3 richer representations): computed from 13:57. D (same-label paired difference) and E (stem-clustered intervals): direction-blind sensitivity analyses specified after their motivating result was known, computed 13:48 and logged 13:52---not preregistered (dated correction in the frozen prereg record). F: signed plug-in value (Table~\ref{tab:signed}); logged 16:59, computed 17:03; post-freeze, not preregistered.
\end{itemize}

\textbf{Analysis environment.}
Seeded analysis (base seed 20260901; per-stage seeds in the frozen analysis record), 10,000 task-clustered bootstrap draws in every preregistered interval (2,000 in the extension analyses, App.~\ref{app:ext}), Python 3.10.8 with NumPy 1.26.4 for the least-squares solve. The statistical analysis pipeline made no model API calls and collected no new rollouts; the endpoint-map audit of App.~\ref{app:audit2} used \$1.15 of hosted-API calls to two LLM labelers.
Total downstream wall time was 786 s.

\section{Endpoint method map and audit protocol}\label{app:map}

\textbf{Map.}
The map records the HTTP method attached to each AppWorld v0.1.3.post1 endpoint decorator; GET is read and POST/PUT/PATCH/DELETE write.
The map was extracted once from the AppWorld v0.1.3.post1 endpoint decorators and is recorded with the analysis (aliases in Table~\ref{tab:alias}); of the 119 functions observed in the cohort's pre-boundary function sets, 18 are explicit aliases.
Method counts: 76 GET, 35 POST, 5 DELETE, 3 PATCH; 0 unmapped.
A boundary is \emph{post-write} under the original label if any function in its pre-boundary function set is non-GET, and \emph{substantive-write} under the post hoc label if any non-GET function's name suffix is outside the session list of App.~\ref{app:prereg}.

\begin{table}[htbp]
\centering
\footnotesize
\caption{The 18 explicit aliases in the endpoint method map.}
\label{tab:alias}
\begin{tabular}{@{}llll@{}}
\toprule
Function & Method & Function & Method \\
\midrule
\texttt{file\_system.list\_directory} & GET & \texttt{supervisor.show\_account} & GET \\
\texttt{file\_system.read} & GET & \texttt{supervisor.show\_account\_credentials} & GET \\
\texttt{file\_system.read\_file} & GET & \texttt{supervisor.show\_account\_profile} & GET \\
\texttt{file\_system.show\_file\_metadata} & GET & \texttt{supervisor.show\_account\_usernames} & GET \\
\texttt{phone.get\_contact} & GET & \texttt{supervisor.show\_phone\_contacts} & GET \\
\texttt{phone.set\_access\_token} & POST & \texttt{venmo.show\_api\_descriptions} & GET \\
\texttt{simple\_note.show\_api\_descriptions} & GET & \texttt{venmo.show\_balance} & GET \\
\texttt{spotify.get\_music\_queue} & GET & \texttt{venmo.show\_friends} & GET \\
\texttt{spotify.show\_music\_player} & GET & \texttt{venmo.show\_transaction\_aggregate\_stats} & GET \\
\bottomrule
\end{tabular}
\end{table}

\textbf{Check of the map.}
A separately encoded semantic map---an agent-authored table encoded in the analysis---gives a per-endpoint judgement for the 43 non-GET functions observed in the cohort (function name plus AppWorld API semantics, one reason each) and a read-verb rule for the remaining 76 (name begins with \texttt{show\_}, \texttt{search\_}, \texttt{get\_}, \texttt{read} or \texttt{list\_}); the table agrees with the decorator method on 43/43.
A 50-boundary comparison (25 per original cell, seed 20260908) follows mechanically from that table and the pre-boundary function sets and is not a second audit; an earlier comparison that derived its labels from the map itself was discarded as circular.
A further audit by two labelers of different model families, blind to the map, is reported in App.~\ref{app:audit2}.

\subsection{Blind two-labeler audit}\label{app:audit2}
In a second, map-blind audit under a protocol frozen beforehand, two LLM labelers of different model families were given the READ/WRITE and SESSION/SUBSTANTIVE definitions but not the map. They were called through a hosted API; the frozen protocol specified temperature 0 for both, but one model's endpoint rejects that setting, so it ran at its default temperature (a logged deviation) and the other at temperature 0. Each labeler saw each of the 119 endpoint names with the one-line AppWorld description where one could be extracted from the corpus's documentation calls (101 of 119; the 18 alias names have none) and never the method map, and answered READ or WRITE and, for writes, SESSION or SUBSTANTIVE.
Both labelers agree with the decorator map on 119 of 119 endpoints (43 of 43 non-GET, 76 of 76 GET), with each other on 119 of 119, and with the session-suffix rule on 43 of 43 writes (11 session, 32 substantive); there were no parse failures.
Replacing the map by either labeler's calls moves 0 of 590 boundaries under the original split (494/96) and 0 under the post hoc split (126/464).
API cost \$1.15; per-call cost records are kept in the analysis record.

\begin{table}[htbp]
\centering
\footnotesize
\caption{Session endpoints among the 494 boundaries labelled post-write under the original label: boundaries whose pre-boundary function set contains the function.}
\label{tab:auth}
\begin{tabular}{@{}lr@{}}
\toprule
Function & Boundaries \\
\midrule
\texttt{spotify.login} & 200 \\
\texttt{phone.login} & 146 \\
\texttt{venmo.login} & 143 \\
\texttt{file\_system.login} & 99 \\
\texttt{simple\_note.login} & 73 \\
\texttt{gmail.login} & 4 \\
\texttt{spotify.send\_password\_reset\_code} & 4 \\
\texttt{venmo.logout} & 2 \\
\texttt{venmo.send\_verification\_code} & 2 \\
\texttt{phone.set\_access\_token} & 2 \\
\texttt{spotify.reset\_password} & 1 \\
\bottomrule
\end{tabular}
\end{table}

\section{Cell estimates, exploratory contrasts and regressions}\label{app:cells}

All intervals are 95\% task-clustered percentile bootstrap intervals with 10,000 draws.
Under the original label the cells are read-only (96 boundaries, 60 tasks) and any prior write (494, 135); under the post hoc label, no substantive write (464, 132) and substantive write (126, 64).

\begin{table}[htbp]
\centering
\footnotesize
\setlength{\tabcolsep}{4pt}
\caption{Per-cell paired deltas by horizon and channel, original label.}
\label{tab:cells-orig}
\begin{tabular}{@{}llrll@{}}
\toprule
Cell & $k$ & \multicolumn{1}{c}{dU} & \multicolumn{1}{c}{dE} & \multicolumn{1}{c}{dR} \\
\midrule
Read-only & 1 & $+$0.135 [$+$0.071, $+$0.205] & $+$0.053 [$+$0.010, $+$0.101] & $+$0.086 [$+$0.026, $+$0.148] \\
Read-only & 3 & $+$0.398 [$+$0.266, $+$0.533] & $+$0.130 [$+$0.061, $+$0.212] & $+$0.272 [$+$0.160, $+$0.384] \\
Read-only & 5 & $+$0.681 [$+$0.504, $+$0.859] & $+$0.179 [$+$0.107, $+$0.266] & $+$0.516 [$+$0.370, $+$0.665] \\
Any prior write & 1 & $+$0.130 [$+$0.097, $+$0.163] & $+$0.119 [$+$0.096, $+$0.144] & $+$0.021 [$-$0.008, $+$0.048] \\
Any prior write & 3 & $+$0.335 [$+$0.269, $+$0.405] & $+$0.229 [$+$0.188, $+$0.270] & $+$0.140 [$+$0.088, $+$0.194] \\
Any prior write & 5 & $+$0.476 [$+$0.384, $+$0.570] & $+$0.299 [$+$0.246, $+$0.353] & $+$0.231 [$+$0.157, $+$0.305] \\
\bottomrule
\end{tabular}
\end{table}

\begin{table}[htbp]
\centering
\footnotesize
\setlength{\tabcolsep}{4pt}
\caption{Per-cell paired deltas by horizon and channel, post hoc label.}
\label{tab:cells-v2}
\begin{tabular}{@{}llrll@{}}
\toprule
Cell & $k$ & \multicolumn{1}{c}{dU} & \multicolumn{1}{c}{dE} & \multicolumn{1}{c}{dR} \\
\midrule
No substantive write & 1 & $+$0.153 [$+$0.119, $+$0.188] & $+$0.110 [$+$0.087, $+$0.134] & $+$0.049 [$+$0.023, $+$0.075] \\
No substantive write & 3 & $+$0.393 [$+$0.326, $+$0.464] & $+$0.224 [$+$0.186, $+$0.264] & $+$0.197 [$+$0.145, $+$0.250] \\
No substantive write & 5 & $+$0.566 [$+$0.478, $+$0.658] & $+$0.291 [$+$0.243, $+$0.343] & $+$0.316 [$+$0.248, $+$0.385] \\
Substantive write & 1 & $+$0.048 [$-$0.004, $+$0.099] & $+$0.104 [$+$0.053, $+$0.160] & $-$0.034 [$-$0.096, $+$0.030] \\
Substantive write & 3 & $+$0.172 [$+$0.048, $+$0.291] & $+$0.171 [$+$0.077, $+$0.266] & $+$0.031 [$-$0.065, $+$0.123] \\
Substantive write & 5 & $+$0.301 [$+$0.122, $+$0.472] & $+$0.241 [$+$0.116, $+$0.366] & $+$0.136 [$-$0.013, $+$0.274] \\
\bottomrule
\end{tabular}
\end{table}

\textbf{Contrasts.} (Only the original-label all-boundaries contrast was prespecified; the first-compaction restriction is amendment G2$'$(d), and the substantive-write contrasts use the post hoc label.)
Original label, dU@3, post-write minus read-only: all boundaries $-0.062$ [$-0.209$, $+0.079$], half-width 0.144 (590 boundaries, 140 tasks, 55 tasks with both cells); first compactions only $+0.069$ [$-0.086$, $+0.225$], half-width 0.156 (275 boundaries, 42 tasks with both cells).
Post hoc label, substantive-write minus no-substantive-write: all boundaries $-0.221$ [$-0.359$, $-0.093$], half-width 0.133 (56 tasks with both cells); first compactions only $-0.389$ [$-0.546$, $-0.234$], half-width 0.156, with only 5 tasks containing both cells (18 substantive-write first compactions out of 275).

\begin{table}[htbp]
\centering
\footnotesize
\setlength{\tabcolsep}{4pt}
\caption{Exploratory horizon $\times$ channel contrasts (not prespecified; 590 boundaries, 140 tasks). Left: original label, post-write minus read-only. Right: post hoc label, substantive-write minus no-substantive-write.}
\label{tab:explore}
\begin{tabular}{@{}llll@{}}
\toprule
$k$ & Channel & Original label & Post hoc label \\
\midrule
1 & dU & $-$0.005 [$-$0.078, $+$0.066] & $-$0.105 [$-$0.167, $-$0.044] \\
1 & dE & $+$0.067 [$+$0.013, $+$0.117] & $-$0.006 [$-$0.062, $+$0.053] \\
1 & dR & $-$0.066 [$-$0.135, $+$0.002] & $-$0.083 [$-$0.150, $-$0.012] \\
3 & dU & $-$0.062 [$-$0.207, $+$0.079] & $-$0.221 [$-$0.359, $-$0.092] \\
3 & dE & $+$0.099 [$+$0.003, $+$0.182] & $-$0.053 [$-$0.153, $+$0.047] \\
3 & dR & $-$0.132 [$-$0.254, $-$0.010] & $-$0.166 [$-$0.274, $-$0.066] \\
5 & dU & $-$0.205 [$-$0.405, $-$0.009] & $-$0.265 [$-$0.461, $-$0.081] \\
5 & dE & $+$0.120 [$+$0.018, $+$0.209] & $-$0.050 [$-$0.183, $+$0.083] \\
5 & dR & $-$0.285 [$-$0.457, $-$0.118] & $-$0.180 [$-$0.341, $-$0.033] \\
\bottomrule
\end{tabular}
\end{table}

\textbf{Regressions.}
Outcome: per-boundary dU@3.
Uncontrolled: intercept and label.
Controlled: intercept, label, log prefix steps, a first-compaction indicator and 139 task indicators; ordinary least squares; task-clustered bootstrap intervals.
Original label, post-write coefficient: uncontrolled $-0.062$ [$-0.207$, $+0.078$]; controlled $-0.027$ [$-0.216$, $+0.174$].
Post hoc label, substantive-write coefficient: uncontrolled $-0.221$ [$-0.358$, $-0.091$]; controlled $-0.004$ [$-0.193$, $+0.176$].
The uncontrolled OLS coefficient on a binary regressor is the difference of cell means, so the three printed $-0.221$ figures are one estimate ($-0.2209$); their intervals differ only because each table draws its own 10,000 task resamples (seeds 20260924, 20260930, 20260932); the same holds for the three $-0.062$ intervals of the original-label contrast.

\textbf{Exposure by arm.}
A rollout's $L$ is its number of recorded steps (equal to its valid primary actions; a rollout stops at five actions, when the environment reports the task complete, or at a step with no valid call). Boundary-weighted over the 590 boundaries (2,320 rollouts per arm; five zero-step rollouts enter with $L = 0$): valid actions through $k = 3$, PRE 2.768 / POST 2.761; completed by $k = 3$ (terminated with task complete at a step $\le 3$), PRE 0.221 / POST 0.199; through $k = 5$, 4.249 / 4.299 and 0.368 / 0.328 (the release's own whole-rollout columns). The exposure imbalance is small; burden is reported as a total effect that includes failure to continue; a reach-$k$ sensitivity is future work.

\section{The first-compaction confound}\label{app:first}

\IfFileExists{figures/fig2-label-construct.pdf}{\begin{figure}[tbp]
\centering
\includegraphics[width=0.62\linewidth]{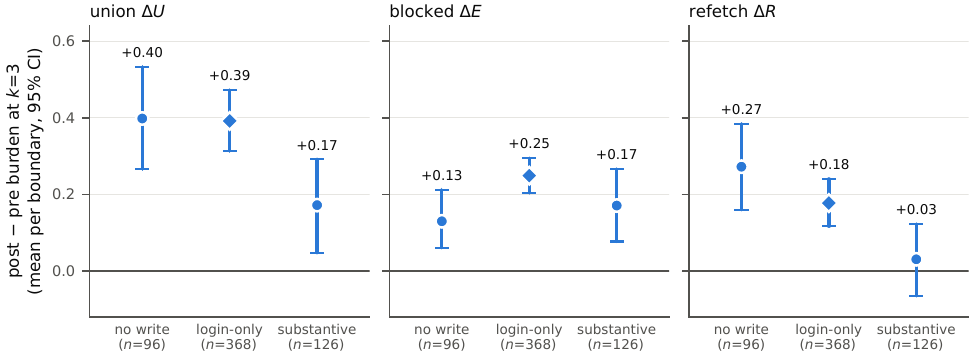}
\caption{Paired deltas at $k=3$ (dU, dE, dR; mean per boundary, 95\% task-clustered intervals) for the three pre-boundary constructs of Table~\ref{tab:label}: no write (96 boundaries; mean prefix 7.9 steps), login-only write (368; 15.1), substantive write (126; 22.1). The no-write and substantive-write intervals are those of Table~\ref{tab:label}; the login-only interval is derived (seed 20260950). The login-only group is not a cell of either frozen label; the figure is descriptive.}
\label{fig:construct}
\end{figure}}{}
The cohort has 275 first and 315 later compactions.
Under the original label the read-only cell is 78/96 first compactions (81\%) and the write cell 197/494; under the post hoc label 257 of the 464 no-substantive-write boundaries and 18 of the 126 substantive-write boundaries are first compactions.
Mean compaction index is 1.22 for no-write, 1.89 for login-only and 3.14 for substantive-write boundaries (Table~\ref{tab:label}).
Any label that correlates with prefix length therefore also correlates with whether the boundary is a first compaction, and the repository's shipped curves already separate first from later boundaries and stratify by compaction index \citep{trace_software_2026}; first and later boundaries are also sampled differently (five versus three per arm) \citep{min2026trace}.
This is why the controlled regressions of App.~\ref{app:cells} include a first-compaction indicator and why the first-compaction-only contrasts are reported: under the original label the first-compaction-only contrast is $+0.069$ [$-0.086$, $+0.225$], and under the post hoc label the corresponding contrast rests on 5 tasks that contain both cells.
In the controlled regressions the first-compaction indicator itself carries $+0.274$ [$+0.084$, $+0.474$] under the original label and $+0.273$ [$+0.091$, $+0.466$] under the post hoc label (log prefix steps $-0.103$ [$-0.368$, $+0.178$] and $-0.113$ [$-0.366$, $+0.158$]).

\section{Trigger policies: definitions, folds, frontiers and sensitivity}\label{app:policy}

\textbf{Observables.}
\begin{itemize}\itemsep1pt\raggedright
\item \texttt{post\_write\_prefix}: the original label (the post hoc runs use the post hoc substantive-write label \texttt{substantive\_write\_prefix});
\item \texttt{log\_prefix\_steps} $=\log(1+\textrm{prefix steps})$;
\item \texttt{recent\_error\_count}: the number of the last three prefix observations that begin with the environment's failure contract;
\item \texttt{last\_observation\_chars}: the length of the last prefix observation;
\item \texttt{compression\_index}: 1 for a trajectory's first compaction, 2 for its second, and so on;
\item \texttt{n\_historical\_exact}: the number of distinct exact call signatures before the boundary.
\end{itemize}
The stump and the depth-2 tree choose among all six; the logistic uses the label and \texttt{recent\_error\_count}.

\textbf{Fitting.}
Trees: Gini impurity, at most 25 candidate thresholds per feature, no split of fewer than 20 rows.
Logistic: features standardised on the training fold, full-batch gradient descent (1,500 iterations, step 0.05, L2 penalty 0.01 on non-intercept weights).
Folds: the 140 task identifiers are shuffled once with a fixed seed and assigned round-robin to five folds; a boundary is scored by the model fitted on the other four folds.
Frontier: boundaries are sorted by out-of-fold risk score; a threshold allows compaction for every boundary with score strictly below it; the operating point is the threshold that maximises the avoided fraction among all thresholds retaining at least 80\% of boundaries; intervals are task-clustered, conditional on the fitted out-of-fold scores and the realized labels (10,000 draws; seeds 20261112, 20261124, 20261128 for stump, depth-2, logistic); a nested bootstrap that also resamples the 3--5 replays within each boundary (Table~\ref{tab:nested}; 2,000 draws, seed 20260960, scores fixed, no refit) leaves the widths essentially unchanged but shifts the intervals: the count advantages over the random-rule expectation become $+0.044$ [$+0.006$, $+0.065$] (logistic), $+0.037$ [$+0.010$, $+0.069$] (depth-2), $+0.016$ [$-0.005$, $+0.051$] (stump) and $+0.047$ [$+0.025$, $+0.089$] (boosted), and no burden-mass advantage then excludes zero among the four cells of Table~\ref{tab:nested} (the 144-feature cell's does; App.~\ref{app:ext3}); the count-advantage half-widths grow by less than 0.004 (full widths by up to 0.007; the avoided-fraction half-widths change by at most 0.006, Table~\ref{tab:nested}), and we read the nested family as a sensitivity analysis, not a second confidence family.

\begin{table}[htbp]
\centering
\footnotesize
\setlength{\tabcolsep}{4pt}
\caption{Nested (replays within boundary, then tasks) versus task-only bootstrap intervals for the avoided fraction at $\ge 80\%$ retained; half-widths in parentheses; last column: burden-mass advantage over the random-rule expectation under the nested bootstrap. Under replay resampling 11.5\% of boundaries change sign label and the resampled positive rate is 0.533 against the realized 0.583.}
\label{tab:nested}
\begin{tabular}{@{}lrlll@{}}
\toprule
Policy & Avoided & Nested 95\% CI & Task-only 95\% CI & Mass advantage (nested) \\
\midrule
Stump & 0.108 & [0.084, 0.239] (0.077) & [0.083, 0.238] (0.078) & $+0.001$ [$-0.032$, $+0.033$] \\
Depth-2 tree & 0.195 & [0.112, 0.251] (0.069) & [0.096, 0.246] (0.075) & $+0.042$ [$-0.003$, $+0.074$] \\
Logistic (2 features) & 0.206 & [0.146, 0.250] (0.052) & [0.156, 0.256] (0.050) & $-0.001$ [$-0.041$, $+0.031$] \\
17 summaries (boosted) & 0.247 & [0.223, 0.288] (0.032) & [0.217, 0.277] (0.030) & $+0.038$ [$-0.005$, $+0.083$] \\
\bottomrule
\end{tabular}
\end{table}

\begin{table}[htbp]
\centering
\footnotesize
\setlength{\tabcolsep}{4pt}
\caption{Signed plug-in value at the $\ge 80\%$ operating point (amendment F, specified after review). $V$ = burden avoided by keeping PRE on the blocked set $B$, with boundaries where POST had less burden counted against the policy ($\Sigma^{+}$, $\Sigma^{-}$, number negative); $E_{\mathrm{rand}} = (|B|/590)\,\Sigma_{\mathrm{all}}$ with $\Sigma_{\mathrm{all}}\,\mathrm{dU@3} = 203.9$ over 590 boundaries (mean 0.346). Intervals: task-clustered, 2,000 draws (seed 20260964), and nested task $\times$ replay (seed 20260965), scores and thresholds fixed. The positive-part version ($\max(0,\mathrm{dU})$) reproduces the burden-mass column of Table~\ref{tab:frontier-full} and the mass advantages of App.~\ref{app:ext} in point estimate; these intervals hold the operating threshold fixed, whereas Table~\ref{tab:ext2c}'s reselect it in every draw, so a borderline cell's zero crossing can differ between the two.}
\label{tab:signed}
\begin{tabular}{@{}lrlrl@{}}
\toprule
Policy & $|B|$ (share) & $V$ ($\Sigma^{+}$, $\Sigma^{-}$; neg.) & $E_{\mathrm{rand}}$ & $V-E_{\mathrm{rand}}$ [task 95\% CI] \\
\midrule
Stump & 54 (0.092) & 22.00 (25.00, $-3.00$; 8) & 18.67 & $+3.33$ [$-4.46$, $+11.67$] \\
Depth-2 tree & 93 (0.158) & 49.67 (54.27, $-4.60$; 12) & 32.15 & $+17.52$ [$+4.62$, $+29.69$] \\
Logistic (frozen) & 96 (0.163) & 38.20 (44.00, $-5.80$; 11) & 33.18 & $+5.02$ [$-6.25$, $+16.82$] \\
Boosted, 17 summaries & 118 (0.200) & 59.27 (64.73, $-5.47$; 12) & 40.79 & $+18.48$ [$+5.52$, $+31.11$] \\
Boosted, 144 features & 118 (0.200) & 74.07 (76.80, $-2.73$; 8) & 40.79 & $+33.28$ [$+19.14$, $+47.78$] \\
Logistic, 144 features & 118 (0.200) & 78.53 (82.87, $-4.33$; 9) & 40.79 & $+37.75$ [$+23.72$, $+52.78$] \\
Logistic, summaries + difficulty & 118 (0.200) & 66.00 (72.13, $-6.13$; 11) & 40.79 & $+25.21$ [$+11.52$, $+38.34$] \\
\bottomrule
\end{tabular}

\medskip
\begin{tabular}{@{}llll@{}}
\toprule
Policy & $V-E_{\mathrm{rand}}$ [nested 95\% CI] & $V/|B|$ [task 95\% CI] & [nested] \\
\midrule
Stump & [$-5.59$, $+12.51$] & 0.407 [0.257, 0.550] & [0.237, 0.575] \\
Depth-2 tree & [$+3.80$, $+31.46$] & 0.534 [0.378, 0.676] & [0.370, 0.696] \\
Logistic (frozen) & [$-6.90$, $+18.16$] & 0.398 [0.266, 0.531] & [0.262, 0.550] \\
Boosted, 17 summaries & [$+5.09$, $+32.46$] & 0.502 [0.376, 0.624] & [0.369, 0.639] \\
Boosted, 144 features & [$+18.90$, $+48.81$] & 0.628 [0.502, 0.753] & [0.500, 0.765] \\
Logistic, 144 features & [$+22.77$, $+55.28$] & 0.666 [0.535, 0.804] & [0.531, 0.820] \\
Logistic, summaries + difficulty & [$+10.61$, $+40.94$] & 0.559 [0.431, 0.680] & [0.426, 0.702] \\
\bottomrule
\end{tabular}
\end{table}
\textbf{Signed plug-in value.} The signed plug-in value exceeds its random-blocking expectation by $+5.0$ [$-6.3$, $+16.8$] burden-steps for the frozen logistic (96 blocked; not clear of zero), $+18.5$ [$+5.5$, $+31.1$] for the boosted seventeen-summary trigger and $+33.3$ [$+19.1$, $+47.8$] for the 144-feature one (118 blocked each; 0.40--0.63 per blocked boundary against a mean of 0.35; post hoc; Table~\ref{tab:signed}).

Amendment F extended the nested analysis from five to seven policies: the nested positive-part interval excludes zero for three of the seven (boosted-144 [$+8.3$, $+35.2$], logistic-144 [$+13.4$, $+41.7$], logistic on summaries + difficulty [$+4.5$, $+28.5$] burden-steps); the depth-2 tree's touches zero ([$-0.04$, $+21.3$]); the task-only positive-part interval clears zero in four of seven (the boosted seventeen-summary model's, [$-0.6$, $+21.0$], does not), the signed value in five of seven under both bootstraps.
$\Delta$ compares against the fixed expectation $1 - \mathrm{Ret.}$; a realized uniformly random rule blocking 54 / 93 / 96 / 118 of 590 boundaries has a hypergeometric standard deviation of 0.010 / 0.013 / 0.013 / 0.014 in the avoided fraction, so the stump's $+0.016$ is about 1.6 such deviations and the frozen logistic's $+0.044$ about 3.4.
In Fig.~\ref{fig:frontier} the depth-2, logistic and 144-feature markers are offset horizontally by 0.008 and the markers at 0.80 overprint, their intervals ($\pm 0.03$) lying inside the marker.
The frontier pools out-of-fold scores from the five fold models and is therefore a mixture of five fitted policies (a deployable stump has a single operating point; the stump's interior frontier points exist only because the fold models have different leaf rates); per-fold AUROC, mean (range), beside the pooled value: stump 0.618 (0.570--0.669) against 0.592, depth-2 0.633 (0.578--0.701) against 0.619, logistic 0.582 (0.525--0.642) against 0.547; the pooled scores take 10 / 20 / 24 distinct values and the fold positive rates range 0.514--0.641.
Advantage over the random rule by burden mass at the operating points of Table~\ref{tab:frontier}: logistic $-0.001$ [$-0.034$, $+0.035$], stump $+0.001$ [$-0.028$, $+0.034$], depth-2 $+0.042$ [$+0.007$, $+0.079$], boosted $+0.038$ [$+0.003$, $+0.086$].

\textbf{Fitted models (descriptive).}
The stump splits on \texttt{n\_historical\_exact} $\le 15.5$ in four folds and on \texttt{compression\_index} $\le 1.5$ in one, the shorter-history leaf having the higher positive-burden rate in all five (0.68--0.71 against 0.40--0.47).
The depth-2 tree's short-history $\times$ short-last-observation leaf (distinct prior calls $\le 15.5$, or compaction index $\le 1.5$ in one fold, and last observation $\le$ 50--60 characters) has a positive-burden rate of 0.89--0.93 in four of five folds on the rows the fold model was fitted on ($n$ = 28--36 per fold; 0.71 in the fifth); these are in-sample leaf purities of outcome-selected splits, reported descriptively. The release caps observations at 200 characters and 464 of 590 boundaries sit at the cap; the 74 boundaries at $\le 60$ characters are ones whose last call returned almost nothing---an empty result list or a one-line status or lookup reply---mostly first compactions after short histories; their positive-burden share is 0.73 against 0.58 overall.
The depth-2 tree refines that split mostly with \texttt{last\_observation\_chars} on the shorter-history side and \texttt{log\_prefix\_steps} on the longer-history side, in one fold each with \texttt{compression\_index} and \texttt{n\_historical\_exact}.
The logistic coefficients on the standardised features are negative for both features in every fold under both labellings.
This records what the policies used, not a mechanism.

\begin{table}[htbp]
\centering
\footnotesize
\setlength{\tabcolsep}{4pt}
\caption{Task-held-out AUROC \citep{hanley1982meaning} by horizon and channel for the target $1[\Delta>0]$ of that channel. Left block: original label; right block: post hoc label. Each policy is refit out-of-fold to that horizon $\times$ channel target under the same task-held-out folds; only the $k=3$ union cells coincide with the frozen dU@3 policies.}
\label{tab:auroc}
\begin{tabular}{@{}llrrrrrr@{}}
\toprule
 & & \multicolumn{3}{c}{Original label} & \multicolumn{3}{c}{Post hoc label} \\
$k$ & Channel & Stump & Depth-2 & Logistic & Stump & Depth-2 & Logistic \\
\midrule
1 & dU & 0.546 & 0.577 & 0.467 & 0.546 & 0.572 & 0.539 \\
1 & dE & 0.490 & 0.567 & 0.503 & 0.490 & 0.567 & 0.475 \\
1 & dR & 0.572 & 0.602 & 0.507 & 0.572 & 0.602 & 0.515 \\
3 & dU & 0.592 & 0.619 & 0.547 & 0.592 & 0.628 & 0.560 \\
3 & dE & 0.468 & 0.466 & 0.489 & 0.468 & 0.466 & 0.438 \\
3 & dR & 0.600 & 0.635 & 0.532 & 0.600 & 0.635 & 0.533 \\
5 & dU & 0.580 & 0.597 & 0.540 & 0.585 & 0.608 & 0.551 \\
5 & dE & 0.474 & 0.491 & 0.404 & 0.474 & 0.491 & 0.450 \\
5 & dR & 0.610 & 0.604 & 0.547 & 0.610 & 0.625 & 0.570 \\
\bottomrule
\end{tabular}
\end{table}

The horizon $\times$ channel grid of Table~\ref{tab:auroc} and the exploratory contrasts of App.~\ref{app:cells} are descriptive and not adjusted for multiplicity.

\textbf{Stem-clustered sensitivity} (48 task-id stems as clusters, 2,000 draws; task-clustered value in parentheses; these intervals re-estimate the policies' uncertainty, not their fitting, which stays task-held-out).
Placement contrast [$-0.238$, $+0.100$] ([$-0.209$, $+0.079$]); avoided at $\ge 80\%$---stump [0.085, 0.239], depth-2 [0.099, 0.252], logistic [0.146, 0.260], boosted [0.217, 0.278]; count advantage---stump [$-0.003$, $+0.050$], depth-2 [$+0.013$, $+0.063$], logistic [$+0.013$, $+0.071$], boosted [$+0.018$, $+0.079$]; mass advantage---stump [$-0.032$, $+0.038$], depth-2 [$+0.008$, $+0.080$], logistic [$-0.041$, $+0.039$], boosted [$-0.002$, $+0.088$]; replicate AUROC [0.689, 0.747]; same-label difference boosted [$+0.031$, $+0.117$], depth-2 [$+0.052$, $+0.153$], logistic [$+0.146$, $+0.262$]; boosted AUROC [0.607, 0.709].
Stem clustering widens the intervals modestly (placement half-width 0.144 $\to$ 0.169, the rest by at most 0.01) and changes no sign conclusion except the boosted mass advantage's lower bound ($+0.003 \to -0.002$), consistent with the nested result.

\textbf{Sensitivity.}
Leave-one-task-stem-out values are means of within-stem AUROCs (48 stems, 1--35 boundaries per stem; 47 finite values, one stem single-class) and are a different estimator from the pooled AUROC of Table~\ref{tab:auroc}, not comparable to it: 0.630 (stump), 0.670 (depth-2), 0.578 (logistic) under the original label and 0.630 / 0.649 / 0.567 under the post hoc label.
Restricting to boundaries with $|\mathrm{dU@3}|\ge 0.2$ ($n=462$; $|\mathrm{dU@3}|\ge 0.2$ within $10^{-9}$, 453 under exact comparison): 0.606 / 0.607 / 0.535 (original) and 0.606 / 0.632 / 0.546 (post hoc).
Post hoc label operating points at $\ge 80\%$ retained: stump unchanged; depth-2 0.235 [0.108, 0.253] at 0.805 retained; logistic 0.116 [0.066, 0.181] at 0.886 retained.

\begin{table}[htbp]
\centering
\footnotesize
\setlength{\tabcolsep}{4pt}
\caption{Full out-of-fold frontiers, original label: every threshold at which the allowed set changes. Retained = fraction of the 590 boundaries allowed to compact; avoided = fraction of the 344 positive-burden boundaries blocked; mass = fraction of positive burden mass blocked.}
\label{tab:frontier-full}
\begin{tabular}{@{}lrrrr@{}}
\toprule
Policy & Allowed & Retained & Avoided & Mass \\
\midrule
Stump & 0 & 0.000 & 1.000 & 1.000 \\
 & 59 & 0.100 & 0.904 & 0.885 \\
 & 104 & 0.176 & 0.837 & 0.818 \\
 & 144 & 0.244 & 0.802 & 0.788 \\
 & 212 & 0.359 & 0.698 & 0.682 \\
 & 258 & 0.437 & 0.657 & 0.639 \\
 & 330 & 0.559 & 0.506 & 0.498 \\
 & 401 & 0.680 & 0.363 & 0.339 \\
 & 471 & 0.798 & 0.230 & 0.206 \\
 & 536 & 0.908 & 0.108 & 0.092 \\
 & 590 & 1.000 & 0.000 & 0.000 \\
\midrule
Depth-2 & 0 & 0.000 & 1.000 & 1.000 \\
 & 15 & 0.025 & 0.977 & 0.972 \\
 & 33 & 0.056 & 0.948 & 0.938 \\
 & 34 & 0.058 & 0.945 & 0.935 \\
 & 57 & 0.097 & 0.927 & 0.919 \\
 & 72 & 0.122 & 0.904 & 0.894 \\
 & 90 & 0.153 & 0.898 & 0.888 \\
 & 120 & 0.203 & 0.855 & 0.849 \\
 & 173 & 0.293 & 0.773 & 0.769 \\
 & 214 & 0.363 & 0.706 & 0.688 \\
 & 236 & 0.400 & 0.677 & 0.664 \\
 & 259 & 0.439 & 0.654 & 0.637 \\
 & 327 & 0.554 & 0.515 & 0.505 \\
 & 392 & 0.664 & 0.387 & 0.369 \\
 & 449 & 0.761 & 0.285 & 0.275 \\
 & 497 & 0.842 & 0.195 & 0.200 \\
 & 561 & 0.951 & 0.076 & 0.088 \\
 & 565 & 0.958 & 0.064 & 0.079 \\
 & 571 & 0.968 & 0.047 & 0.062 \\
 & 577 & 0.978 & 0.032 & 0.038 \\
 & 590 & 1.000 & 0.000 & 0.000 \\
\midrule
Logistic & 0 & 0.000 & 1.000 & 1.000 \\
 & 1 & 0.002 & 1.000 & 1.000 \\
 & 2 & 0.003 & 0.997 & 0.998 \\
 & 4 & 0.007 & 0.991 & 0.994 \\
 & 5 & 0.008 & 0.991 & 0.994 \\
 & 7 & 0.012 & 0.991 & 0.994 \\
 & 21 & 0.036 & 0.968 & 0.967 \\
 & 38 & 0.064 & 0.939 & 0.931 \\
 & 57 & 0.097 & 0.919 & 0.916 \\
 & 64 & 0.108 & 0.913 & 0.911 \\
 & 73 & 0.124 & 0.910 & 0.910 \\
 & 168 & 0.285 & 0.738 & 0.704 \\
 & 257 & 0.436 & 0.576 & 0.538 \\
 & 342 & 0.580 & 0.433 & 0.398 \\
 & 429 & 0.727 & 0.302 & 0.259 \\
 & 494 & 0.837 & 0.206 & 0.162 \\
 & 496 & 0.841 & 0.201 & 0.156 \\
 & 497 & 0.842 & 0.198 & 0.154 \\
 & 498 & 0.844 & 0.195 & 0.146 \\
 & 499 & 0.846 & 0.195 & 0.146 \\
 & 517 & 0.876 & 0.148 & 0.116 \\
 & 531 & 0.900 & 0.116 & 0.090 \\
 & 550 & 0.932 & 0.076 & 0.055 \\
 & 571 & 0.968 & 0.035 & 0.028 \\
 & 590 & 1.000 & 0.000 & 0.000 \\
\bottomrule
\end{tabular}
\end{table}

\begin{figure}[htbp]
\centering
\includegraphics[width=\textwidth]{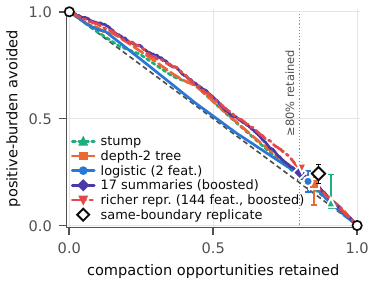}
\caption{Figure~\ref{fig:frontier} reproduced at full width (same data and legend).}
\label{fig:frontier-full}
\end{figure}

\section{Extension analyses: replicate yardstick and engineered-summary trigger}\label{app:ext}

EXT-1 and EXT-2 were frozen, with their interpretation rules, before any output was computed (App.~\ref{app:prereg}); no seed, feature list, hyperparameter, fold scheme or target was changed afterwards; amendments A--C (App.~\ref{app:prereg}) were frozen before their own computation; D, E and F are direction-blind post-freeze analyses specified after their motivating result was known and are not preregistered (dated correction in the frozen prereg record).

\textbf{EXT-1, replicate yardstick.} Splitting each boundary's PRE and POST rollouts into disjoint halves (200 random splits), one half's paired delta predicts the sign of the other's with AUROC 0.69--0.72 at $k=3$ depending on which half supplies the label (0.72 [0.69, 0.75] under the frozen extension protocol's assignment; at later boundaries the smaller half is a single rollout per arm; first 0.74, later 0.68); the half--half correlation is 0.40, the Spearman--Brown step-up gives a nominal full-sample reliability of 0.57 [0.50, 0.62] (the halves are unequal, so this is an extrapolation), and the estimated task-level share of variance is near zero (intraclass correlation 0.01 [$-0.05$, 0.08]; task stem 0.04, point only).
Per rollout, union burden at $k$ is the number of the first $k$ steps that are blocked or refetch; per-boundary mean(POST)\,$-$\,mean(PRE) is required to reproduce the released dU@$k$ for all 590 boundaries before anything else runs.
For 200 random splits, each boundary's PRE samples and POST samples are partitioned into two disjoint halves (2/3 at first compactions, 1/2 at later ones) and a half-A and half-B delta are formed.
Statistics, averaged over splits with 5th--95th percentiles and a task-clustered bootstrap interval on a fixed 20-split subset: Pearson and Spearman correlation between the halves (all / first / later); the \emph{replicate AUROC} of the half-B delta as a score for $1[\textrm{half-A delta}>0]$; sign agreement and Cohen's $\kappa$ on the three-class sign; the Spearman--Brown step-up \citep{spearman1910correlation,brown1910some}, reported with the caveat that it assumes equal parallel halves and ours are unequal; and one-way intraclass correlations of per-rollout burden across boundaries for each arm, and of the per-boundary delta by task and by task stem.
Fixed interpretation rule: the history-based held-out AUROC is reported beside the replicate AUROC; overlapping intervals are read as ``within the range set by the boundary's own replicate'', a replicate interval entirely above as ``fall short of what a replicate of the same boundary predicts''.
The replicate AUROC is a yardstick, not a strict ceiling (a half-sample score against a half-sample label); its own noise floor includes the sampling of three to five rollouts per arm and inference nondeterminism, which persists at temperature zero and across serving configurations \citep{atil2024nondeterminism,yuan2025numerical}.
The split is asymmetric: the smaller half (2 of 5 rollouts per arm at first compactions, 1 of 3 at later ones) supplies the label (positive rate 0.431 against 0.583 realized) and the larger half the score; the mirrored assignment gives 0.69 [0.66, 0.72] at $k=3$ (supplementary, not a preregistered quantity).
On the frontier axes (half-B delta as risk score for the half-A label, operating point at $\ge 80\%$ retained under the rule of App.~\ref{app:policy}), the yardstick avoids 0.243 [0.198, 0.286] at 0.866 retained at $k=3$, $+0.109$ [$+0.082$, $+0.134$] above the random rule (its operating point cannot be placed nearer 0.80 because a 2--3-rollout delta takes few distinct values, so comparisons should be made on the advantage over the random rule, not on the raw avoided fraction); at $k=1$ it avoids 0.355 ($+0.223$ [$+0.180$, $+0.267$]) and at $k=5$ 0.311 ($+0.124$ [$+0.099$, $+0.144$]); at $k=3$ within first compactions 0.241 ($+0.091$ [$+0.062$, $+0.112$]) and within later ones 0.247 ($+0.126$ [$+0.083$, $+0.183$]).
Same-label paired differences (replicate $-$ trigger, both against the half-A label; 2,000 task draws): boosted $+0.073$ [$+0.037$, $+0.111$] (replicate 0.72 against the half-sample label, trigger 0.64 against the same label); depth-2 $+0.101$ [$+0.057$, $+0.146$]; logistic $+0.204$ [$+0.155$, $+0.253$] (the logistic's scores discriminate the half-sample label at 0.513 against 0.547 for the realized label). The mixed-label comparison (App.~\ref{app:prereg}, amendment B; replicate against the half-sample label, triggers against the realized label: $+0.061$ [$+0.008$, $+0.113$]; $+0.098$ [$+0.048$, $+0.152$]; $+0.170$ [$+0.108$, $+0.230$]) is retained as context only.
Results are in Table~\ref{tab:ext1}; the interpretation sentence fixed by the rule above is the one in \S\ref{sec:trigger}.
One-way intraclass correlations at $k=3$: per-rollout union burden across boundaries, PRE arm 0.60 [0.54, 0.66], POST arm 0.49 [0.44, 0.53]; per-boundary delta by task 0.01 [$-$0.05, 0.08], by task stem 0.04 (point only: the task-resampling interval is not valid for stem grouping).

\begin{table}[htbp]
\centering
\footnotesize
\setlength{\tabcolsep}{4pt}
\caption{EXT-1 split-half replicate statistics by horizon and subset (all 590 boundaries; 275 first compactions; 315 later). Means over 200 random splits; intervals are task-clustered bootstrap intervals (2,000 draws) of the mean over a fixed subset of 20 splits. Replicate AUROC = half-B delta as a score for $1[\textrm{half-A delta}>0]$; $r_{\mathrm{full}}$ = Spearman--Brown step-up of the half--half Pearson correlation; sign agreement treats zero as its own class.}
\label{tab:ext1}
\begin{tabular}{@{}llrlrlrr@{}}
\toprule
$k$ & Subset & $n$ & Replicate AUROC [95\% CI] & Pearson & $r_{\mathrm{full}}$ (Spearman--Brown) [95\% CI] & Sign agr. & $\kappa$ \\
\midrule
1 & all & 590 & 0.726 [0.691, 0.760] & 0.407 & 0.578 [0.514, 0.631] & 0.630 & 0.331 \\
1 & first & 275 & 0.765 [0.725, 0.807] & 0.550 & 0.710 [0.656, 0.763] & 0.661 & 0.393 \\
1 & later & 315 & 0.673 [0.608, 0.726] & 0.320 & 0.484 [0.372, 0.568] & 0.602 & 0.263 \\
3 & all & 590 & 0.718 [0.686, 0.746] & 0.397 & 0.568 [0.505, 0.624] & 0.551 & 0.282 \\
3 & first & 275 & 0.742 [0.699, 0.783] & 0.509 & 0.674 [0.615, 0.737] & 0.609 & 0.296 \\
3 & later & 315 & 0.683 [0.634, 0.725] & 0.331 & 0.496 [0.384, 0.580] & 0.500 & 0.237 \\
5 & all & 590 & 0.723 [0.688, 0.745] & 0.416 & 0.587 [0.523, 0.638] & 0.573 & 0.303 \\
5 & first & 275 & 0.760 [0.711, 0.795] & 0.512 & 0.676 [0.610, 0.735] & 0.654 & 0.346 \\
5 & later & 315 & 0.677 [0.623, 0.713] & 0.364 & 0.532 [0.426, 0.605] & 0.502 & 0.238 \\
\bottomrule
\end{tabular}
\end{table}

\textbf{EXT-2, engineered-summary trigger.}
Seventeen pre-boundary features fixed in advance (log prefix steps, compaction index, first-compaction flag, distinct exact and function calls, prefix steps minus distinct exact calls (a repeat proxy that is negative when a step issues several calls), distinct apps, documentation-lookup calls and their share, session-call count, the original write label and the post hoc substantive-write label, count of substantive-write functions, recent errors, log last-observation length, any failure in the three-observation tail, presence of the task-completion call), plus task difficulty as a separately reported group (a three-level one-hot, so the H+T cells carry 20 columns).
Models with fixed hyperparameters: L2 logistic regression ($C=1$; lbfgs solver, at most 5,000 iterations) and a gradient-boosted classifier (depth 3, learning rate 0.05, 200 iterations, minimum 20 rows per leaf, L2 regularisation 1.0, no early stopping).
Five-fold cross-validation grouped by task stem (48 stems, stricter than the task-held-out folds of \S\ref{sec:trigger}), 20 repetitions with shuffled group assignment; task-grouped folds as a sensitivity.
Primary target $1[\mathrm{dU@3}>0]$; sensitivity targets at $k=1$ and $k=5$ and on the blocked and refetch channels; a regression variant scored by held-out Spearman correlation.
Metrics: held-out AUROC (mean and range over repetitions, task-clustered bootstrap interval), average precision, and the operating point at $\ge 80\%$ retained under exactly the rule of App.~\ref{app:policy} with the random-rule expectation beside it; out-of-fold permutation importance reported descriptively.
``Best'' is the model $\times$ feature set with the highest mean held-out AUROC on the primary target, named before any sensitivity target is examined; it is one of four eligible cells and its interval is not selection-adjusted.
At the operating point the largest advantage is the logistic model on the summaries plus difficulty ($+0.073$ [$+0.045$, $+0.100$]), not the cell with the highest AUROC; the two post hoc substantive-write comparators sit below the random-rule expectation (Table~\ref{tab:ext2c}).
Results for the primary target under stem-grouped folds are in Tables~\ref{tab:ext2}--\ref{tab:ext2c}.
Two estimators are reported: the AUROC of the repetition-averaged out-of-fold scores, whose task-clustered bootstrap interval (2,000 draws) is the protocol's and which the body uses everywhere (best cell 0.66 [0.61, 0.70]; with task difficulty 0.65 [0.61, 0.70]; frozen feature pairs 0.48--0.51 [0.43, 0.56]); and the mean over the 20 repetitions of the per-repetition pooled AUROC (best cell 0.65, range 0.62--0.66; frozen pairs 0.54--0.56, supplementary interval [0.50, 0.60]).
For the seventeen-feature models the two agree within 0.012; for the two-feature models (10--24 distinct score values) the per-repetition pooled value differs from the repetition-averaged value, and since the per-repetition scores are not released the difference cannot be decomposed further; the frozen G4 numbers of \S\ref{sec:trigger}, from a single task-held-out split, are unaffected.
For the best model (boosted classifier, seventeen history features), on the repetition-averaged-score estimator, the sensitivity targets give held-out AUROC 0.61 at $k=1$, 0.61 at $k=5$, 0.56 on the blocked channel and 0.65 on the refetch channel; task-grouped folds give 0.66; the regression variant reaches held-out Spearman correlation 0.30 [0.21, 0.38] with dU@3.
Out-of-fold permutation importance for the boosted model with task difficulty, reported descriptively as mean AUROC drop: number of distinct prior exact calls 0.045, compaction index 0.039, post hoc substantive-write label 0.018, number of distinct prior functions 0.017, log last-observation length 0.012.
Two implementation notes: H17 (task-completion call present in the exact set) is constant zero over the 590 boundaries and drops out; and for one boundary the POST sample carries a different observation tail, so the first error-free PRE rollout's tail was used.

\begin{table}[htbp]
\centering
\footnotesize
\setlength{\tabcolsep}{3pt}
\caption{EXT-2 held-out discrimination, primary target $1[\mathrm{dU@3}>0]$, five-fold cross-validation grouped by task stem, 20 repetitions. First AUROC column: AUROC of the repetition-averaged out-of-fold scores with its task-clustered bootstrap 95\% CI (the protocol's interval); second: mean over repetitions of the per-repetition pooled AUROC (min--max); AP = average precision (mean over repetitions). H = the seventeen history features; T = task difficulty; the last four rows are the two frozen feature pairs of \S\ref{sec:trigger} under these folds (``errors'' = recent errors).}
\label{tab:ext2}
\begin{tabular}{@{}lllll@{}}
\toprule
Model & Features & AUROC of rep.-avg.\ scores [95\% CI] & Per-rep.\ mean (min--max) & AP \\
\midrule
Boosted & H (17) & 0.657 [0.608, 0.703] & 0.645 (0.623--0.659) & 0.699 \\
L2 logistic & H (17) & 0.629 [0.575, 0.680] & 0.624 (0.606--0.641) & 0.698 \\
Boosted & H+T (20) & 0.654 [0.605, 0.700] & 0.642 (0.621--0.656) & 0.698 \\
L2 logistic & H+T (20) & 0.626 [0.574, 0.676] & 0.621 (0.599--0.638) & 0.701 \\
L2 logistic & has-written + errors & 0.482 [0.429, 0.532] & 0.544 (0.526--0.565) & 0.617 \\
Boosted & has-written + errors & 0.483 [0.429, 0.535] & 0.543 (0.524--0.563) & 0.616 \\
L2 logistic & post hoc subst.-write + errors & 0.512 [0.458, 0.561] & 0.563 (0.536--0.584) & 0.615 \\
Boosted & post hoc subst.-write + errors & 0.502 [0.449, 0.554] & 0.558 (0.542--0.581) & 0.613 \\
\bottomrule
\end{tabular}
\end{table}

\begin{table}[htbp]
\centering
\footnotesize
\setlength{\tabcolsep}{3pt}
\caption{EXT-2 operating point at $\ge 80\%$ retained under the rule of App.~\ref{app:policy}, same models and folds as Table~\ref{tab:ext2}; the random-rule expectation is $1-$retained.}
\label{tab:ext2b}
\begin{tabular}{@{}llrll@{}}
\toprule
Model & Features & Ret. & Avoided [95\% CI] & Mass [95\% CI] \\
\midrule
Boosted & H (17) & 0.800 & 0.247 [0.217, 0.277] & 0.238 [0.201, 0.284] \\
L2 logistic & H (17) & 0.800 & 0.262 [0.233, 0.286] & 0.248 [0.211, 0.286] \\
Boosted & H+T (20) & 0.800 & 0.250 [0.219, 0.273] & 0.249 [0.208, 0.287] \\
L2 logistic & H+T (20) & 0.800 & 0.273 [0.243, 0.299] & 0.266 [0.227, 0.303] \\
L2 logistic & has-written + errors & 0.829 & 0.209 [0.180, 0.255] & 0.164 [0.134, 0.214] \\
Boosted & has-written + errors & 0.805 & 0.227 [0.183, 0.256] & 0.179 [0.134, 0.213] \\
L2 logistic & post hoc subst.-write + errors & 0.802 & 0.177 [0.104, 0.205] & 0.166 [0.093, 0.202] \\
Boosted & post hoc subst.-write + errors & 0.814 & 0.142 [0.111, 0.183] & 0.130 [0.091, 0.179] \\
\bottomrule
\end{tabular}
\end{table}

\begin{table}[htbp]
\centering
\footnotesize
\setlength{\tabcolsep}{4pt}
\caption{EXT-2 paired advantage over the random-rule expectation at the operating points of Table~\ref{tab:ext2b}, by count and by burden mass (task-clustered, 2,000 draws).}
\label{tab:ext2c}
\begin{tabular}{@{}llll@{}}
\toprule
Model & Features & $\Delta$ count [95\% CI] & $\Delta$ mass [95\% CI] \\
\midrule
Boosted & H (17) & $+$0.047 [$+$0.019, $+$0.079] & $+$0.038 [$+$0.003, $+$0.086] \\
L2 logistic & H (17) & $+$0.062 [$+$0.035, $+$0.088] & $+$0.048 [$+$0.012, $+$0.088] \\
Boosted & H+T (20) & $+$0.050 [$+$0.021, $+$0.075] & $+$0.049 [$+$0.010, $+$0.088] \\
L2 logistic & H+T (20) & $+$0.073 [$+$0.045, $+$0.100] & $+$0.066 [$+$0.029, $+$0.105] \\
L2 logistic & has-written + errors & $+$0.038 [$-$0.002, $+$0.065] & $-$0.007 [$-$0.051, $+$0.026] \\
Boosted & has-written + errors & $+$0.032 [$-$0.004, $+$0.067] & $-$0.016 [$-$0.053, $+$0.028] \\
L2 logistic & post hoc subst.-write + errors & $-$0.021 [$-$0.052, $+$0.010] & $-$0.032 [$-$0.069, $+$0.008] \\
Boosted & post hoc subst.-write + errors & $-$0.044 [$-$0.072, $-$0.011] & $-$0.057 [$-$0.095, $-$0.015] \\
\bottomrule
\end{tabular}
\end{table}

\clearpage
\subsection{Richer representations (144--8,336 features; EXT-3)}\label{app:ext3}

Adding one indicator per prior function and app to the summaries (144 features) gives held-out AUROC 0.66 [0.61, 0.71] and $+0.067$ [$+0.040$, $+0.092$] over the random-rule expectation at 80\% retained; hashed observation text or exact calls did not improve held-out AUROC point estimates (0.61--0.65); the 144-feature model's burden-mass advantage stays clear of zero under replay resampling ($+0.083$, nested [$+0.026$, $+0.123$]; task-only [$+0.040$, $+0.125$]), and the same-label gap to the replicate is $+0.06$ [$+0.02$, $+0.10$].

\begin{table}[htbp]
\centering
\footnotesize
\setlength{\tabcolsep}{4pt}
\caption{EXT-3 held-out discrimination, primary target $1[\mathrm{dU@3}>0]$, stem-grouped five-fold cross-validation, 20 repetitions; same learners, folds, repetitions, target, metrics and bootstrap draws as EXT-2. AUROC: repetition-averaged-score estimator [95\% CI] and per-repetition mean (min--max); AP = average precision.}
\label{tab:ext3}
\begin{tabular}{@{}llrllr@{}}
\toprule
Learner & Repr. & Features & AUROC [95\% CI] & Per-rep.\ mean (min--max) & AP \\
\midrule
L2 logistic & R1 & 144 & 0.654 [0.606, 0.698] & 0.644 (0.622--0.661) & 0.712 \\
L2 logistic & R2 & 4,240 & 0.606 [0.555, 0.655] & 0.594 (0.569--0.621) & 0.672 \\
L2 logistic & R3 & 8,336 & 0.612 [0.561, 0.660] & 0.601 (0.569--0.628) & 0.680 \\
Boosted & R1 & 144 & 0.663 [0.612, 0.708] & 0.651 (0.635--0.667) & 0.709 \\
Boosted & R2 & 4,240 & 0.642 [0.592, 0.686] & 0.628 (0.607--0.648) & 0.688 \\
Boosted & R3 & 8,336 & 0.651 [0.603, 0.695] & 0.636 (0.621--0.655) & 0.695 \\
\bottomrule
\end{tabular}
\end{table}

\begin{table}[htbp]
\centering
\footnotesize
\setlength{\tabcolsep}{4pt}
\caption{EXT-3 operating point at 0.800 retained, same cells as Table~\ref{tab:ext3}; $\Delta$ = paired advantage over the random-rule expectation, by count and by burden mass (intervals paired with EXT-2's).}
\label{tab:ext3b}
\begin{tabular}{@{}llrlll@{}}
\toprule
Learner & Repr. & Features & Avoided [95\% CI] & $\Delta$ count [95\% CI] & $\Delta$ mass [95\% CI] \\
\midrule
L2 logistic & R1 & 144 & 0.273 [0.243, 0.296] & $+0.073$ [$+0.045$, $+0.098$] & $+0.105$ [$+0.053$, $+0.144$] \\
L2 logistic & R2 & 4,240 & 0.253 [0.221, 0.276] & $+0.053$ [$+0.023$, $+0.077$] & $+0.083$ [$+0.035$, $+0.123$] \\
L2 logistic & R3 & 8,336 & 0.253 [0.225, 0.279] & $+0.053$ [$+0.026$, $+0.080$] & $+0.081$ [$+0.042$, $+0.124$] \\
Boosted & R1 & 144 & 0.267 [0.238, 0.290] & $+0.067$ [$+0.040$, $+0.092$] & $+0.083$ [$+0.040$, $+0.125$] \\
Boosted & R2 & 4,240 & 0.262 [0.231, 0.284] & $+0.062$ [$+0.033$, $+0.085$] & $+0.096$ [$+0.050$, $+0.131$] \\
Boosted & R3 & 8,336 & 0.262 [0.234, 0.288] & $+0.062$ [$+0.036$, $+0.089$] & $+0.093$ [$+0.054$, $+0.135$] \\
\bottomrule
\end{tabular}
\end{table}

R1 = the seventeen summaries plus one binary indicator per function observed in the cohort (119) and per app (8); R2 = R1 plus a hashed binary bag of word tokens from the three tail observations (4,096 buckets); R3 = R2 plus a hashed bag of exact call signatures (4,096 buckets); same learners, folds, repetitions, target, metrics and bootstrap draws as EXT-2 (intervals paired with EXT-2's).
Best cell by the fixed rule: boosted trees on R1; it is one of six cells and its interval is not selection-adjusted.
Its sensitivity targets (repetition-averaged AUROC): $k=1$ 0.59 [0.55, 0.64]; $k=5$ 0.62 [0.58, 0.66]; blocked 0.56 [0.51, 0.61]; refetch 0.66 [0.62, 0.70]; task-grouped folds 0.66 [0.62, 0.71].
Permutation importance (AUROC drop): distinct prior calls 0.048, distinct prior functions 0.022, post hoc substantive-write label 0.016, compaction index 0.013, last-observation length 0.013; the largest identity feature is the spotify-app indicator at 0.007; 101 of 144 features never change the AUROC when permuted.
Robustness of the best cell: the nested task $\times$ replay bootstrap gives avoided [0.245, 0.304], $\Delta$ count [$+0.048$, $+0.105$] and $\Delta$ mass [$+0.026$, $+0.123$] (task-only [0.238, 0.290], [$+0.040$, $+0.092$], [$+0.040$, $+0.125$]); unlike the summary-only cell, its mass advantage stays clear of zero under replay resampling---17 trigger cells examined on burden mass; task-only intervals exclude zero upward in 11 (and downward in one, the boosted post hoc substantive-write model); the nested design was run for 7 (five under amendment A, all seven under amendment F) and excludes zero in 3 (the two 144-feature models and the logistic on summaries + difficulty). Same-label replicate $-$ trigger difference $+0.061$ [$+0.025$, $+0.097$] (stem-clustered [$+0.021$, $+0.104$]); stem-clustered AUROC [0.615, 0.714], avoided [0.240, 0.289], $\Delta$ count [$+0.042$, $+0.090$], $\Delta$ mass [$+0.042$, $+0.125$]. By AUROC the boosted cell is best; the logistic on the same 144 features has the larger operating-point advantage ($+0.073$ count, $+0.105$ mass).

\section{LLM-use disclosure}\label{app:llm}

The outcomes in this paper are the source corpus's recorded blocked and refetch flags under its environment's error contract. We made no model call that judged an outcome or generated a rollout; the subject model is the corpus's own.
Two instruments were LLMs: the encoded semantic-map check of App.~\ref{app:map} and the two map-blind endpoint labelers of App.~\ref{app:audit2} (hosted-API calls costing \$1.15); the statistical pipeline itself made no model calls.
Beyond those instruments, models assisted with the analysis code and with drafting the manuscript, directed and reviewed by the authors, who are responsible for every claim, number and measurement.
Reported statistics were checked against the frozen analysis artifacts, and cited works against their sources.

\end{document}